%% file: example_paper.tex
\documentclass{article}

\usepackage{bostyle}
\usepackage{microtype}
\usepackage{graphicx}
\usepackage{subfigure}
\usepackage{booktabs} % for professional tables
\usepackage{multirow}
\usepackage{xcolor}
\usepackage[T1]{fontenc}

\definecolor{MyDarkRed}{rgb}{0.69,0.0,0.1098}

\usepackage{hyperref}

\usepackage[accepted]{icml2021}

\icmltitlerunning{Coach-Player Multi-Agent Reinforcement Learning}

\begin{document}

\twocolumn[
\icmltitle{Coach-Player Multi-Agent Reinforcement Learning\\ for Dynamic Team Composition}

% It is OKAY to include author information, even for blind
% submissions: the style file will automatically remove it for you
% unless you've provided the [accepted] option to the icml2021
% package.

% List of affiliations: The first argument should be a (short)
% identifier you will use later to specify author affiliations
% Academic affiliations should list Department, University, City, Region, Country
% Industry affiliations should list Company, City, Region, Country

% You can specify symbols, otherwise they are numbered in order.
% Ideally, you should not use this facility. Affiliations will be numbered
% in order of appearance and this is the preferred way.
\icmlsetsymbol{equal}{*}

\begin{icmlauthorlist}
\icmlauthor{Bo Liu}{ut,equal}
\icmlauthor{Qiang Liu}{ut}
\icmlauthor{Peter Stone}{ut}
\icmlauthor{Animesh Garg}{uot,nv}
\icmlauthor{Yuke Zhu}{ut,nv}
\icmlauthor{Animashree Anandkumar}{nv,ct}

\end{icmlauthorlist}

\icmlaffiliation{ut}{Department of Computer Science, The University of Texas at Austin, Austin, USA}
\icmlaffiliation{uot}{University of Toronto, Toronto, Canada}
\icmlaffiliation{ct}{California Institute of Technology, Pasadena, USA}
\icmlaffiliation{nv}{Nvidia}

\icmlcorrespondingauthor{Bo Liu}{bliu@cs.utexas.edu}

% You may provide any keywords that you
% find helpful for describing your paper; these are used to populate
% the "keywords" metadata in the PDF but will not be shown in the document
\icmlkeywords{Machine Learning, ICML}

\vskip 0.3in
]

% this must go after the closing bracket ] following \twocolumn[ ...

% This command actually creates the footnote in the first column
% listing the affiliations and the copyright notice.
% The command takes one argument, which is text to display at the start of the footnote.
% The \icmlEqualContribution command is standard text for equal contribution.
% Remove it (just {}) if you do not need this facility.

%\printAffiliationsAndNotice{}  % leave blank if no need to mention equal contribution
\printAffiliationsAndNotice{\textsuperscript{*}Work done during an internship at NVIDIA.} % otherwise use the standard text.

\input{content/abstract}

\input{content/intro}
\input{content/background}

\input{content/method}
\input{content/experiment}
\input{content/related}
\input{content/conclusion}

% Acknowledgements should only appear in the accepted version.
%\section*{Acknowledgements}

% In the unusual situation where you want a paper to appear in the
% references without citing it in the main text, use \nocite
%\nocite{langley00}

\bibliography{example_paper}
\bibliographystyle{icml2021}
\appendix
\input{content/appendix}

%%%%%%%%%%%%%%%%%%%%%%%%%%%%%%%%%%%%%%%%%%%%%%%%%%%%%%%%%%%%%%%%%%%%%%%%%%%%%%%
%%%%%%%%%%%%%%%%%%%%%%%%%%%%%%%%%%%%%%%%%%%%%%%%%%%%%%%%%%%%%%%%%%%%%%%%%%%%%%%
% DELETE THIS PART. DO NOT PLACE CONTENT AFTER THE REFERENCES!
%%%%%%%%%%%%%%%%%%%%%%%%%%%%%%%%%%%%%%%%%%%%%%%%%%%%%%%%%%%%%%%%%%%%%%%%%%%%%%%
%%%%%%%%%%%%%%%%%%%%%%%%%%%%%%%%%%%%%%%%%%%%%%%%%%%%%%%%%%%%%%%%%%%%%%%%%%%%%%%
% \appendix
% \section{Do \emph{not} have an appendix here}

% \textbf{\emph{Do not put content after the references.}}
% %
% Put anything that you might normally include after the references in a separate
% supplementary file.

% We recommend that you build supplementary material in a separate document.
% If you must create one PDF and cut it up, please be careful to use a tool that
% doesn't alter the margins, and that doesn't aggressively rewrite the PDF file.
% pdftk usually works fine. 

% \textbf{Please do not use Apple's preview to cut off supplementary material.} In
% previous years it has altered margins, and created headaches at the camera-ready
% stage. 
%%%%%%%%%%%%%%%%%%%%%%%%%%%%%%%%%%%%%%%%%%%%%%%%%%%%%%%%%%%%%%%%%%%%%%%%%%%%%%%
%%%%%%%%%%%%%%%%%%%%%%%%%%%%%%%%%%%%%%%%%%%%%%%%%%%%%%%%%%%%%%%%%%%%%%%%%%%%%%%

\end{document}

\twocolumn[
\icmltitle{Appendix: Coach-Player Multi-Agent Reinforcement Learning\\ for Dynamic Team Composition}

% It is OKAY to include author information, even for blind
% submissions: the style file will automatically remove it for you
% unless you've provided the [accepted] option to the icml2021
% package.

% List of affiliations: The first argument should be a (short)
% identifier you will use later to specify author affiliations
% Academic affiliations should list Department, University, City, Region, Country
% Industry affiliations should list Company, City, Region, Country

% You can specify symbols, otherwise they are numbered in order.
% Ideally, you should not use this facility. Affiliations will be numbered
% in order of appearance and this is the preferred way.
\icmlsetsymbol{equal}{*}

\begin{icmlauthorlist}
\icmlauthor{Aeiau Zzzz}{equal,to}
\icmlauthor{Bauiu C.~Yyyy}{equal,to,goo}
\icmlauthor{Cieua Vvvvv}{goo}
\icmlauthor{Iaesut Saoeu}{ed}
\icmlauthor{Fiuea Rrrr}{to}
\icmlauthor{Tateu H.~Yasehe}{ed,to,goo}
\icmlauthor{Aaoeu Iasoh}{goo}
\icmlauthor{Buiui Eueu}{ed}
\icmlauthor{Aeuia Zzzz}{ed}
\icmlauthor{Bieea C.~Yyyy}{to,goo}
\icmlauthor{Teoau Xxxx}{ed}
\icmlauthor{Eee Pppp}{ed}
\end{icmlauthorlist}

\icmlaffiliation{to}{Department of Computation, University of Torontoland, Torontoland, Canada}
\icmlaffiliation{goo}{Googol ShallowMind, New London, Michigan, USA}
\icmlaffiliation{ed}{School of Computation, University of Edenborrow, Edenborrow, United Kingdom}

\icmlcorrespondingauthor{Cieua Vvvvv}{c.vvvvv@googol.com}
\icmlcorrespondingauthor{Eee Pppp}{ep@eden.co.uk}

% You may provide any keywords that you
% find helpful for describing your paper; these are used to populate
% the "keywords" metadata in the PDF but will not be shown in the document
\icmlkeywords{Machine Learning, ICML}

\vskip 0.3in
]

% this must go after the closing bracket ] following \twocolumn[ ...

% This command actually creates the footnote in the first column
% listing the affiliations and the copyright notice.
% The command takes one argument, which is text to display at the start of the footnote.
% The \icmlEqualContribution command is standard text for equal contribution.
% Remove it (just {}) if you do not need this facility.

%\printAffiliationsAndNotice{}  % leave blank if no need to mention equal contribution
\printAffiliationsAndNotice{\icmlEqualContribution} % otherwise use the standard text.

\appendix
\section{Appendix}
\section*{Derivation of the Variational Objective in Section 3.3}
Denote an agent $a$'s future trajectory from time $t+1$ as $\zeta^{a}_t = (o^a_{t+1}, u^a_{t+1}, o^a_{t+2}, u^a_{t+2}, \dots, o^a_{t+T-1}, u^a_{t+T-1})$, then we have
\[
    \begin{split}
    I(z^a_t ; \zeta^a_t, \bm{s}_t) 
    &= \mathbb{E}_{\bm{s}_t, z^a_t, \zeta^a_t}\bigg[\log\frac{q_\xi(z^a_t | \zeta^a_t, \bm{s}_t)}{p(z^a | \bm{s}_t)}\bigg] \\
    &+ \KL\bigg(p(z^a_t | \zeta^a_t, \bm{s}_t), q_\xi(z^a_t | \zeta^a_t, \bm{s}_t))\bigg)\\
    &\geq \mathbb{E}_{\bm{s}_t, z^a_t, \zeta^a_t}\bigg[\log\frac{q_\xi(z^a_t | \zeta^a_t, \bm{s}_t)}{p(z^a | \bm{s}_t)}\bigg] \\
    &= \mathbb{E}_{\bm{s}_t, z^a_t, \zeta^a_t}\bigg[\log q_\xi(z^a_t | \zeta^a_t, \bm{s}_t)\bigg] + H(z^a_t | \bm{s}_t).
    \end{split}
\]
\section*{Proof of Theorem 1}
\label{sec:proof}
Here we expand the assumptions from Theorem 1 and provide the proof for it. The two assumptions are:
\begin{assumption}
\label{ass:1}
Denote the learned team action-value function as $Q^{\text{tot}}$, the learned coach strategy encoder as $f$ and the true optimal action-value function as $Q^\text{tot}_*$. We assume for any $\bm{\tau}_t, \bm{u}_t, \bm{s}_t, \bm{s}_{\hat{t}}, \bm{c}$,
\begin{equation}
||Q^{\text{tot}}(\bm{\tau}_t, \bm{u}_t, f(\bm{s}_{\hat{t}}); \bm{c}) - Q^\text{tot}_*(\bm{s}_t, \bm{u}_t; \bm{c})||_2 \leq \kappa. \end{equation}
\end{assumption}
\begin{assumption}
\label{ass:2}
Denote the learned individual action-value function as $\{Q^{a_i}\}_{i=1}^{n_a}$, and the particular individual action-value at a state $\bm{s}$ with action $\bm{u}$ as $\{q^{a_i} = Q^{a_i}(s^{a_i}, u^{a_i})\}_{i=1}^{n_a}$. Then we assume unilaterally varying any $q^{a_i}$ to $q'$, i.e. all other $\bm{q}^{-a_i}$ remain the same, will not cause dramatic change of $Q^{\text{tot}}$ if $q'$ stays closely to $q^{a_i}$:
\begin{equation}
    \big| Q^{\text{tot}}(\bm{q}^{-a_i}, q^{a_i}) - Q^{\text{tot}}(\bm{q}^{-a_i}, q')\big| \leq \eta_1 |q^{a_i} - q'|
\end{equation}
and for any agent $a$ and $\forall$ $c^a$, $\tau^a_t$, $u^a_t$, $z^a_1$, $z^a_2$ with proper dimensions,
\begin{equation}
    \big| Q^a(\tau^a_t, u^a_t | z^a_1; c^a) - Q^a(\tau^a_t, u^a_t | z^a_2; c^a)\big| \leq \eta_2 ||z^a_1 - z^a_2||_2.
\end{equation}
\end{assumption}
In other words, assumption~\ref{ass:1} assumes the learned $Q^\text{tot}$ approximates the true optimal $Q^\text{tot}_*$ well \emph{combined with} the learned coach strategy function $f$,\footnote{Note here we only assume $Q^\text{tot}$ is accurate around the predicted strategy by $f$, not for any strategy.} and assumption~\ref{ass:2} assumes the learned team action-value $Q^\text{tot}$ has bounded Lipschitz constant. Next we prove Theorem 1.
\begin{proof}
From assumption 2, it is easy to check that if $||\tilde{z^a_t} - z^a_{\hat{t}}||_2 \leq \beta$ for all $a$, then $|Q^{\text{tot}}(\bm{\tau}_t, \bm{u}_t | \bm{\tilde{z}}_t, \bm{c}) - Q^{\text{tot}}(\bm{\tau}_t, \bm{u}_t | \bm{z}_{\hat{t}}, \bm{c})| \leq n_a\eta_1\eta_2\beta$. For notation convenience, we ignore the superscript of $\text{tot}$ and the condition on $\bm{c}$. For a state $\bm{s}$, denote the action the learned policy take as $\bm{u}^\dagger$, $\bm{u}^\dagger = \argmax_{\bm{u}} Q(\bm{\tau}, \bm{u})$. Similarly we can define $\bm{u}^*$ as the action the optimal $Q_*$ takes and $\bm{\tilde{u}}$ that $\tilde{Q}$ takes. From assumption 1, we know that
\begin{equation}
    \begin{split}
    Q_*(\bm{s}, \bm{u}^\dagger) 
    &\geq Q(\bm{\tau}, \bm{u}^\dagger) - \kappa \\
    &\geq Q(\bm{\tau}, \bm{u}^*) - \kappa \geq Q_*(\bm{s}, \bm{u}^*) - 2\kappa.
    \end{split}
\end{equation}
Therefore taking $\bm{u}^\dagger$ will result in at most $2\kappa$ performance drop at this single step. Similarly, denote $\epsilon_0 = n_a\eta_1\eta_2\beta$, then
\begin{equation}
    \begin{split}
    Q(\bm{\tau}, \bm{\tilde{u}}) 
    &\geq \tilde{Q}(\bm{\tau}, \bm{\tilde{u}}) - \epsilon_0 \\
    &\geq \tilde{Q}(\bm{\tau}, \bm{u}^\dagger) - \epsilon_0 \geq Q(\bm{\tau}, \bm{u}^\dagger) - 2\epsilon_0.
    \end{split}
\end{equation}
Hence $Q_*(\bm{s},\bm{\tilde{u}})  \geq Q_*(\bm{s}, \bm{u}^*) - 2(\epsilon_0 + \kappa)$. Note that this means taking the action $\tilde{\bm{u}}$ in the place of $\bm{u}^*$ at state $\bm{s}$ will result in at most $2(\epsilon_0 + \kappa)$ performance drop. This conclusion generalizes to any step $t$. Therefore, if at each single step the performance is bounded within $2(\epsilon_0 + \kappa)$, then overall the performance is within $2(\epsilon_0 + \kappa)/(1 - \gamma)$.
\end{proof}
\begin{table*}[t]
    \centering
    \begin{tabular}{llc}
    \toprule
    Name & Description & Value \\
    \midrule
      $|\mathcal{D}|$   &  replay buffer size & 100000 \\
      $n_\text{head}$   & number of heads in multi-head attention & 4\\
      $n_\text{thread}$ & number of parallel threads for running the environment & 8\\
      \midrule
      $dh$ & the hidden dimension of all modules & 128\\
      $\gamma$ & the discount factor & 0.99 \\
      $lr$ & learning rate & 0.0003 \\
      & optimizer & RMSprop \\
      $\alpha$ & $\alpha$ value in RMSprop & 0.99\\
      $\epsilon$ & $\epsilon$ value in RMSprop & 0.00001 \\
      $n_\text{batch}$ & batch size & 256 \\
      grad clip & clipping value of gradient & 10\\
      target update frequency & how frequent do we update the target network & 200 updates \\
      $\lambda_1$ & $\lambda_1$ in variational objective & 0.001\\
      $\lambda_2$ & $\lambda_2$ in variational objective & 0.0001\\
      \bottomrule
    \end{tabular}
    \caption{Hyper-parameters in our experiments.}
    \label{tab:hyperparam}
\end{table*}
\section*{Network Architecture}
For all experiments, we use the same network architecture where all intermediate hidden layer have 128 dimensions. Note that this is possible since the only difference is the number of entities, which does not influence our architecture when adopting an attention model. The architecture details follow exactly as in Appendix A of ~\citet{iqbal2020ai}.
\section*{Training Details}
To train the model, we set the max total number of steps to 5 million for the resource collection task, 1 million for the rescue games, and 20 million for the StarCraft micromanagement tasks. For both the resource collection task and the rescue games, we use the exponentially decayed $\epsilon$-greedy algorithm as our exploration policy, starting from $\epsilon_0 = 1.0$ to $\epsilon_n = 0.05$. For the StarCraft micromanagement task, we adopt the setting from~\cite{iqbal2020ai} and decay the $\epsilon$ from 1.0 to 0.05 within the first 50,000 steps and keep $\epsilon$ at 0.05 till the end. We parallel all environments with 8 threads for training.
\section*{Hyper Parameters}
\label{sec:hyper-params}
For both the resource collection task and the rescue games, we use the same set of hyper-parameters. We provide them in Table~\ref{tab:hyperparam}.

\bibliography{example_paper}
\bibliographystyle{icml2021}

%% file: content/abstract.tex
\begin{abstract}
    In real-world multi-agent systems, agents with different capabilities may join or leave without altering the team's overarching goals. Coordinating teams with such \emph{dynamic composition} is challenging: the optimal team strategy varies with the composition. We propose COPA, a coach-player framework to tackle this problem. We assume the coach has a global view of the environment and coordinates the players, who only have partial views, by distributing individual \emph{strategies}. Specifically, we 1) adopt the attention mechanism for both the coach and the players; 2) propose a variational objective to regularize learning; and 3) design an adaptive communication method to let the coach decide when to communicate with the players. We validate our methods on a resource collection task, a rescue game, and the StarCraft micromanagement tasks. We demonstrate zero-shot generalization to new team compositions. Our method achieves comparable or better performance than the setting where all players have a full view of the environment. Moreover, we see that the performance remains high even when the coach communicates as little as 13\% of the time using the adaptive communication strategy. Code is available at \url{https://github.com/Cranial-XIX/marl-copa.git}.
\end{abstract}
\vspace{-15pt}
%It also enforces each agent's policy to explicitly condition on the assigned strategy from the coach.

%% file: content/intro.tex
\section{Introduction}
Cooperative multi-agent reinforcement learning (MARL) is the problem of coordinating a team of agents to perform a shared task. It has broad applications in autonomous vehicle teams~\citep{cao2012overview}, sensor networks~\citep{choi2009distributed}, finance~\citep{lee2007multiagent}, and social science~\citep{leibo2017multi}. Recent works in multi-agent reinforcement learning (MARL) powered by deep neural networks have shed light on solving challenging problems such as playing StarCraft~\citep{rashid2018qmix} and soccer~\citep{kurach2020google}. Among these methods, centralized training with decentralized execution (CTDE) has gained extensive attention since learning in a centralized way enables better cooperation while executing independently makes the system efficient and scalable~\citep{lowe2017multi,foerster2018counterfactual,son2019qtran,mahajan2019maven,wang2020multi}. However, most deep CTDE approaches for cooperative MARL are limited to a fixed number of homogeneous agents.

\begin{figure}[t]
    \centering
    \includegraphics[width=\columnwidth]{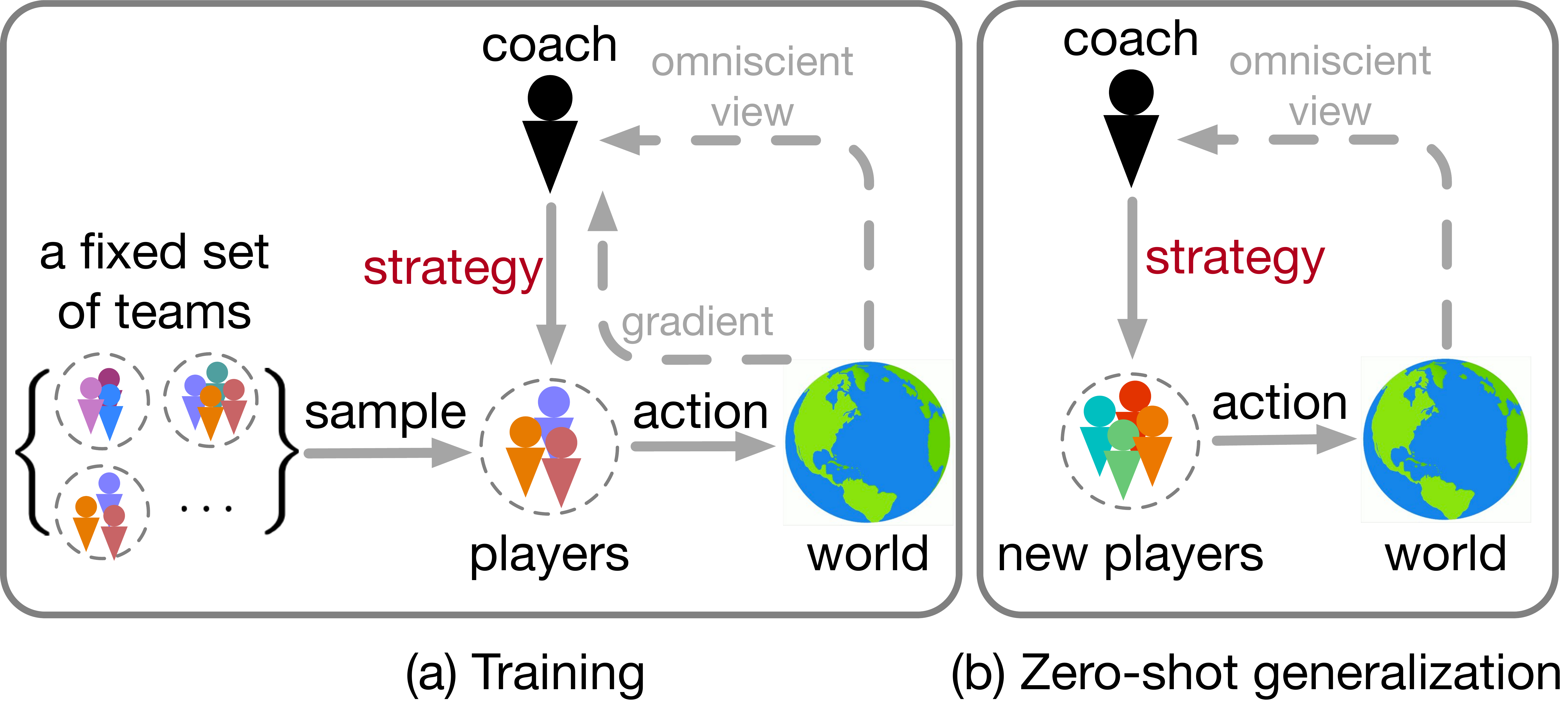}
    \caption{In training, we sample teams from a set of compositions. The coach observes the entire world and coordinates players by broadcasting strategies periodically. In execution, the learned strategy generalizes to new teams.}
    \label{fig:intro}
    \vspace{-15pt}
\end{figure}

Real-world multi-agent tasks, on the other hand, often involve dynamic teams. For example, in a soccer game, a team receiving a red card has one fewer player. In this case, the team may switch to a more defensive strategy. As another example, consider an autonomous vehicle team for delivery. The control over the team depends on how many vehicles we have, how much load each vehicle permits, as well as the delivery destinations. In both examples, the optimal team strategy varies according to the \emph{team composition},\footnote{Team composition is part of an environmental scenario~\citep{de2019multi}, which also includes other environment entities. The formal definition is in Section~\ref{sec:background-problem-formulation}.} i.e., the size of the team and each agent's capability. In these settings, it is computationally prohibitive to re-train the agents for each new team composition. Thus, it is desirable that the model can generalize zero-shot to new team compositions that are unseen during training.

Recently,~\citet{iqbal2020ai} proposed to use the multi-head attention mechanism for modeling a variable number of agents under the CTDE framework.  However, the CTDE constraint can be overly restrictive for complex tasks as each agent only has access to its own decisions and partial environmental observations at test time (see Section~\ref{sec:4.1} for an example where this requirement forbids learning).
The CTDE constraint can be relaxed by introducing communication. To our knowledge, learned communication has only been investigated in homogeneous teams~\cite{foerster2016learning}. On the other hand, prior works on ad hoc teamwork~\cite{barrett2014communicating,grizou2016collaboration,mirsky2020penny} assume a pre-defined communication protocol of which all agents are aware. Moreover, allowing all agents to communicate with each other is too expensive for many scenarios (e.g., battery-powered drones or vehicles). Inspired by real-world sports, we propose to introduce a centralized ``coach" agent who periodically distributes strategic information based on the full view of the environment. We call our method COPA (COach-and-PlAyer).
To our knowledge, this is the first work that investigates learned communication for ad hoc teams.

We introduce a \emph{coach} with an \emph{omniscient} view of the environment, while \emph{player}s only have partial views. We assume that the coach can distribute information to other agents only in limited amounts. We model this communication through a continuous vector, termed as the \emph{strategy} vector, and it is specific to each agent. We design each agent's decision module to incorporate the most recent strategy vector from the coach. Inspired by~\citet{rakelly2019efficient} and \citet{wang2020multi}, we design an additional variational objective to regularize the learning of the strategy. In order to save costs incurred in receiving information from the coach, we design an adaptive policy where the coach communicates with different players only as needed. To train the coach and agents, we sample different teams from a set of team compositions (Figure~\ref{fig:intro}). Recall that the training is centralized under the CTDE framework.\footnote{Rigorously speaking, the players in our method violate the CTDE principle since they occasionally receive global information from the coach. But players still execute independently with local views while they benefit from the centralized learning.} At execution time, the learned policy  generalizes across different team compositions in a zero-shot manner.
% In addition, since dynamic teams with a varying composition can be viewed as a sequence of fixed composition teams, COPA can also generalize to such settings.~\yuke{isn't this just a special case of dynamic team compositions? do we need to highlight this? feels like a subtlety to me}

We design three benchmark environments with dynamic team compositions for evaluating the zero-shot generalization of our method against baselines. They include a resource collection task, a rescue game, and a set of the customized StarCraft micromanagement tasks. All environments are modified such that the team strategy varies with the team composition. We conduct focused ablation studies examining the design choices of COPA on the resource collection task and the rescue games, and further show that COPA applies for more challenging tasks like StarCraft.
Results show comparable or even better performance against methods where players have full observation but no coach. Interestingly, in rescue games, we observe that agents with full views perform worse than agents with partial views, but adding the coach outperforms both. Moreover, with the adaptive communication strategy, we show that the performance remains strong even when the coach communicates as little as 13\% of the time with the player.

{\bf Summary of Contributions: }
\begin{itemize}
    \item We propose a coach-player framework for dynamic team composition of heterogeneous agents.
    \item We introduce a variational objective to regularize the learning of the strategies, leading to faster learning and improved performance, and an adaptive communication strategy to minimize communication from the coach to the agents.
    \item We demonstrate that COPA achieves zero-shot generalization to unseen team compositions. The performance stays strong with a communication frequency as low as 13\%. Moreover, COPA achieves comparable or even better performance than methods where all players have the full views of the environment.
\end{itemize}

%% file: content/background.tex
\section{Background}
In this section, we formally define the learning problem considered in this paper. Then we briefly summarize the idea of value function factorization and a recent improvement that additionally incorporates the attention mechanism, which we leverage in COPA.
\subsection{Problem Formulation} 
\label{sec:background-problem-formulation}
We model the cooperative multi-agent task under the \emph{Decentralized Partially Observable Markov Decision Process} (Dec-POMDP)~\citep{oliehoek2016concise}.
Specifically, we build on the framework of Dec-POMDPs with entities~\citep{de2019multi}, which uses an entity-based knowledge representation. Here, entities include both controllable agents and other environment landmarks. In addition, we extend the representation to allow agents to have individual characteristics, i.e., skill-level, physical condition, etc. Therefore, a Dec-POMDP with characteristic-based entities can be described as a tuple $(\bm{S}, \bm{U}, \bm{O}, P, R, \mathcal{E}, \mathcal{A}, \mathcal{C}, m, \Omega, \rho, \gamma)$. We represent the space of entities by $\mathcal{E}$. For any $e \in \mathcal{E}$, the entity $e$ has a $d_e$ dimensional state representation $s^e \in \mathbb{R}^{d_e}$. The global state is therefore the set $\bm{s} = \{s^e | e\in\mathcal{E}\} \in \bm{S}$. A subset of the entities are controllable agents $a \in \mathcal{A} \subseteq \mathcal{E}$. For both agents and non-agent entities, we differentiate them based on their characteristics $c^e \in \mathcal{C}$.\footnote{$c^e$ is part of $s^e$, but we will explicitly write out $c^e$ in the following for emphasis.} For example, $c^e$ can be a continuous vector that consists of two parts such that only one part can be non-zero. That is, if $e$ is an agent, the first part can represent its skill-level or physical condition, and if $e$ is a non-agent entity, the second part can represent its entity type. A scenario is a multiset of entities $\bm{c} = \{c^e | e \in \mathcal{E}\} \in \Omega$ and possible scenarios are drawn from the distribution $\rho(\bm{c})$. In other words, scenarios are unique up to the composition of the team and that of world entities: a fixed scenario $\bm{c}$ maps to a normal Dec-POMDP with the fixed multiset of entities $\{e | c^e \in \bm{c} \}$.

Given a scenario $\bm{c}$, at each environment step, each agent $a$ can observe a subset of entities specified by an observability function $m: \mathcal{A}\times\mathcal{E} \rightarrow \{0,1\}$, where $m(a, e)$ indicates whether agent $a$ can observe entity $e$.\footnote{An agent always observes itself, i.e., $m(a, a) = 1, \forall a \in \mathcal{A}$.} Therefore, an agent's observation is a set $o^a = \{s^e | m(a, e) = 1\} \in \bm{O}$. Together, at each time step the agents perform a joint action $\bm{u} = \{u^a | a \in \mathcal{A}\} \in \bm{U}$, and the environment will change according to the transition dynamics $P(\bm{s}'|\bm{s},\bm{u}; \bm{c})$. After that, the entire team will receive a single scalar reward $r \sim R(\bm{s}, \bm{u}; \bm{c})$. Starting from an initial state $\bm{s}_0$, the MARL objective is to maximize the discounted cumulative team reward over time:
$G = \mathbb{E}_{\bm{s}_0, \bm{u}_0, \bm{s}_1, \bm{u}_1, \dots} [\sum_{t=0}^\infty \gamma^t r_t]$, where $\gamma$ is the discount factor as is commonly used in RL. Our goal is to learn a team policy that can generalize across different scenarios $\bm{c}$ (different team compositions) and eventually dynamic scenarios (varying team compositions over time).

For optimizing $G$, $Q$-learning is a specific method that makes decisions based on a learned action-value function. The optimal action-value function $Q$ satisfies the Bellman equality:
$Q^{\text{tot}}_*(\bm{s}, \bm{u}; \bm{c}) = r(\bm{s}, \bm{u}; \bm{c}) + \gamma \mathbb{E}_{\bm{s}' \sim P(\cdot | \bm{s}, \bm{u}; \bm{c})}\big[\max_{\bm{u}'} Q^{\text{tot}}_*(\bm{s}', \bm{u}'; \bm{c})\big]$,
where $Q^{\text{tot}}_*$ denote the team's optimal $Q$-value. A common strategy is to adopt function approximation and parameterize the optimal $Q^{\text{tot}}_*$ with parameter $\theta$. Moreover, due to partial observability, the history of observation-action pairs is often encoded with a compact vector representation, i.e., via a recurrent neural network~\citep{medsker1999recurrent}, in place of the state: $Q^\text{tot}_\theta(\bm{\tau}_t, \bm{u}_t; \bm{c}) \approx \mathbb{E}\left[Q^\text{tot}_*(\bm{s}_t, \bm{u}_t; \bm{c})\right]$, where $\bm{\tau} = \{\tau^a | a \in \mathcal{A}\}$ and $\tau^a = (o^a_0, u^a_0, \dots o^a_t)$. In practice, at each time step $t$, the recurrent neural network takes in $(u_{t-1}^a, o_t^a)$ as the new input, where $u_{-1}^a = \bm{0}$ at $t=0$~\citep{zhu2017improving}. Deep $Q$-learning~\citep{mnih2015human} uses deep neural networks to approximate the $Q$ function. In our case, the objective used to train the neural network is:
\begin{equation}
    \label{eq:q-learning}
    \begin{split}
    \mathcal{L}(\theta) &= \mathbb{E}_{(\bm{c}, \bm{\tau}_t, \bm{u}_t, r_t, \bm{\tau}_{t+1}) \sim \mathcal{D}}\bigg[\bigg(r_t +  \\
    &\gamma \max_{\bm{u}'}Q^\text{tot}_{\bar{\theta}}(\bm{\tau}_{t+1}, \bm{u}'; \bm{c}) - Q^\text{tot}_{\theta}(\bm{\tau}_t, \bm{u}_t; \bm{c}) \bigg)^2\bigg].
    \end{split}
\end{equation}
Here, $\mathcal{D}$ is a replay buffer that stores previously generated off-policy data. $Q_{\bar{\theta}}^\text{tot}$ is the target network parameterized by a delayed copy of $\theta$ for stability~\citep{mnih2015human}.

\subsection{Value Function Factorization and Attention QMIX}
Factorizing the action-value function $Q$ into per-agent value functions is a popular approach in centralized training and decentralized execution. Specifically, \citet{rashid2018qmix} propose QMIX, which factorizes $Q^\text{tot}(\bm{\tau}_t, \bm{u})$ into $\{Q^a(\tau^a_t, u^a | a \in \mathcal{A}\}$ and combines them via a mixing network such that $\forall a, \pdv{Q_\text{tot}}{Q^a} \geq 0$. This condition guarantees that individual optimal action $u^a$ is also the best action for the team. As a result, during execution, the mixing network can be removed and agents can act independently according to their own $Q^a$. Attention QMIX (A-QMIX)~\citep{iqbal2020ai} augments the QMIX algorithm with an attention mechanism to deal with an indefinite number of agents/entities. In particular, for each agent, the algorithm applies the multi-head attention (MHA) layer~\citep{vaswani2017attention} to summarize the information of the other entities. This information is used for both encoding the agent's state and adjusting the mixing network. Specifically, the agent's observation $\bm{o}$ is represented by two matrices: the entity state matrix $\bm{X}^\mathcal{E}$ and the observability matrix $\bm{M}$. Assume that in the current scenario $\bm{c}$, there exists $n_e$ entities, $n_a$ of which are the controllable agents, then $\bm{X}^\mathcal{E} \in \mathbb{R}^{n_e\times d_e}$ endows all entities, with the first $n_a$ rows representing the agents. $\bm{M} \in \{0,1\}^{n_a \times n_e}$ is a binary observability mask and $M_{ij} = m(a^i, e^j)$ indicates whether agent $i$ observes entity $j$. $\bm{X}^\mathcal{E}$ is first passed through an encoder, i.e., a single-layer feed-forward network, the output of which we refer to as $\bm{X}$. Denote the $k$-th row of $\bm{X}$ as $h_k$, then for the $i$-th agent, the MHA layer then takes $h_i$ as the query and $\{h_j | M_{ij} = 1\}$ as the keys to compute a latent representation of $a^i$'s observation. For the mixing network, the same MHA layer takes as input $\bm{X}^\mathcal{E}$ and the full observation matrix $\bm{M}^*$, where $\bm{M}^*_{ij} = 1$ if both $e^i$ and $e^j$ exist in the scenario $\bm{c}$, and outputs the encoded global representation for each agent. These encoded representations are then used to generate the mixing network. For more details, We refer readers to Appendix B of the original A-QMIX paper. While A-QMIX in principle applies to the dynamic team composition problem, it is restricted to fully decentralized execution with partial observation. We leverage the attention modules from A-QMIX but additionally investigate how to efficiently take advantage of global information by introducing a coach.

\citet{iqbal2020ai} proposes an extended version of A-QMIX, called Randomized Entity-wise Factorization for Imagined Learning (REFIL), which randomly breaks up the team into two disjoint parts to further decompose the $Q^{\text{tot}}$. The authors demonstrate with the imaginary grouping, REFIL outperforms A-QMIX on a gridworld resource allocation task and a modified StarCraft environment. As we show in the experiment section, we find that REFIL does not always improve over A-QMIX while doubling the computation resource. For this reason, our method is mainly based on A-QMIX, but extending it to REFIL is straightforward.

%% file: content/method.tex
\begin{figure*}[t!]
    \centering
    \includegraphics[width=0.83\textwidth]{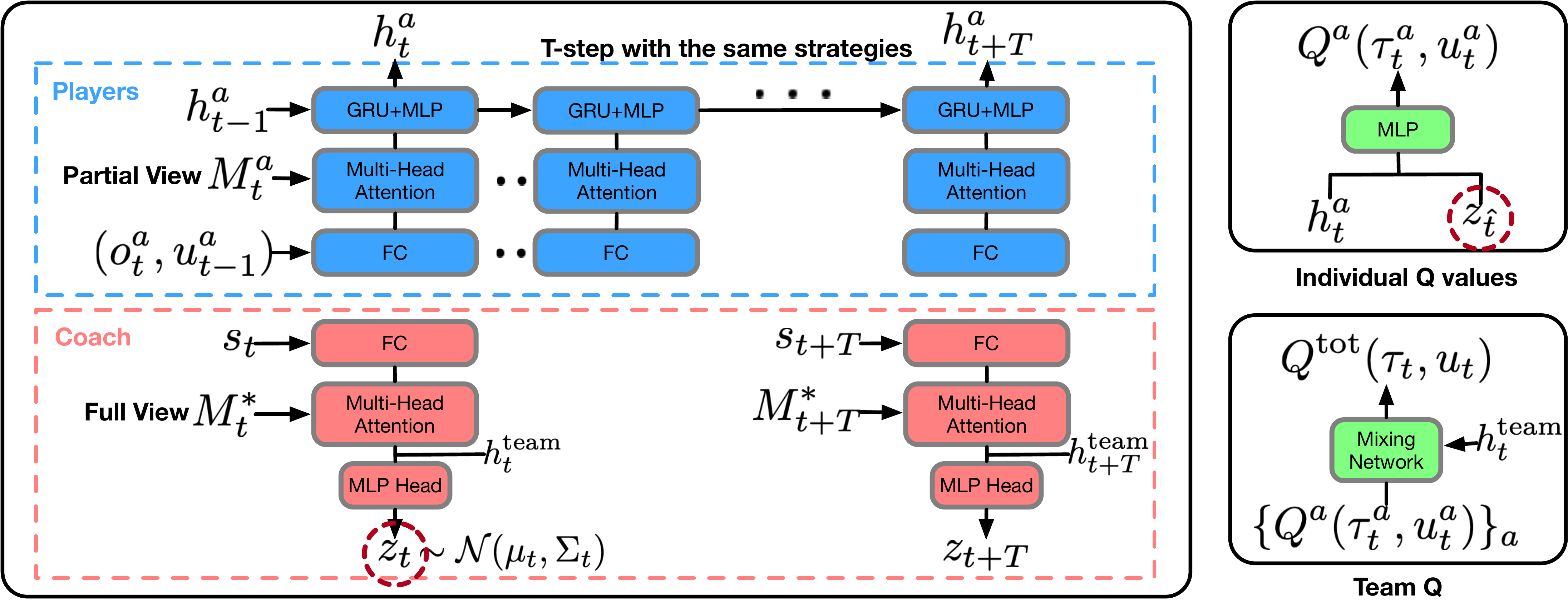}
    \caption{The coach-player network architecture. Here, GRU means gated recurrent unit~\citep{chung2014empirical}; MLP means multi-layer perceptron; and FC means fully connected layer. Both coach and players use multi-head attention to encode information. The coach has an omniscient view while players have partial views. $h^a_t$ encodes agent $a$'s observation history. $h^a_t$ includes the most recent strategy $z_{\hat{t}} = z_{t - t\%T}$ to predict the individual utility $Q^a$ (strategy $z$ is circled in {\color{MyDarkRed}\textbf{red}}). The mixing network uses the hidden layer from the coach, $h^{\text{team}}_t$ to combine all $Q^a$s to predict $Q^\text{tot}$.}
    \label{fig:architecture}
\end{figure*}
\section{Method}
Here we introduce a novel coach-player architecture to incorporate global information for adapting the team-level strategy across different scenarios $\bm{c}$. We first introduce the coach agent that coordinates base agents with global information by broadcasting strategies periodically. Then we present the learning objective and and an additional variational objective to regularize the training. We finish by introducing a method to reduce the broadcast rate and provide analysis to support it.
\subsection{On the importance of global information}
\label{sec:4.1}
As the optimal team strategy varies according to the scenario $\bm{c}$, which includes the team composition, it is important for the team to be aware of any scenario changes as soon as they happen. In an extreme example, consider a multi-agent problem where every agent has its skill-level represented by a real number $c^a \in \mathbb{R}$ and there is a task to complete. For each agent $a$, $u^a \in \{0, 1\}$ indicates whether $a$ chooses to perform the task. The reward is defined as $R(\bm{u}; \bm{c}) = \max_a c^a \cdot u^a + 1 - \sum_a u^a$. In other words, the reward is proportional to the skill-level of the agent who performs it and the team got penalized if more than 1 agent choose to perform the task. If the underlying scenario $\bm{c}$ is fixed, even if all agents are unaware of others' capabilities, it is still possible for the team to gradually figure out the optimal strategy. By contrast, when $\bm{c}$ is subject to change, i.e. agents with different $c$ can join or leave, even if we allow agents to communicate via a network, the information that a particular agent joins or leaves generally takes $d$ time steps to propagate where $d$ is the longest shortest communication path from that agent to any other agents. Therefore, we can see that knowing the global information is not only beneficial but sometimes also necessary for real-time coordination. This motivates the introduction of the coach agent.
\subsection{Coach and players}
We introduce a coach agent and provide it with omniscient observations. To preserve efficiency as in the decentralized setting, we limit the coach agent to only distribute information via a continuous vector $z^a \in \mathbb{R}^{d_z}$ ($d_z$ is the dimension of strategy) to agent $a$, which we call the strategy, once every $T$ time steps. $T$ is the communication interval. The team strategy is therefore represented as $\bm{z} = \{z^a | a \in \mathcal{A}\}$. Strategies are generated via a function $f$ parameterized by $\phi$. Specifically, we assume $z^a \sim \mathcal{N}(\mu^a, \Sigma^a)$, and $(\bm{\mu}=\{\mu^a | a \in \mathcal{A}\}, \bm{\Sigma}=\{\Sigma_a | a \in \mathcal{A}\}) = f_\phi(\bm{s}; \bm{c})$.

For the $T$ steps after receiving $z^a$, agent $a$ will act conditioned on $z^a$. Specifically, within an episode, at time $t_k \in \{v|v\equiv 0~(\mathrm{mod}~T)\}$, the coach observes the global state $\bm{s}_{t_k}$ and  computes and distributes the strategies $\bm{z}_{t_k}$ to all agents. From time $t \in [t_k, t_k+T-1]$, any agent $a$ will act according to its individual action-value $Q^a(\tau^a_t, \cdot \mid z^a_{t_k}; c^a)$.

Denote $\hat{t} = \max \{v|v\equiv 0~(\mathrm{mod}~T)~\text{and}~v \leq t\}$, the most recent time step when the coach distributed strategies. The mean square Bellman error objective in \eqref{eq:q-learning} becomes
\[
    \begin{split}
    &\mathcal{L}_\text{RL}(\theta, \phi) = \mathbb{E}_{(\bm{c}, \bm{\tau}_t, \bm{u}_t, r_t, \bm{s}_{\hat{t}}, \bm{s}_{\hat{t+1}})\sim \mathcal{D}} \bigg[\bigg(r_t +\\
    &\gamma\max_{\bm{u}'}Q^\text{tot}_{\bar{\theta}}(\bm{\tau}_{t+1}, \bm{u}'| \bm{z}_{\hat{t+1}}; \bm{c}) - Q^\text{tot}_\theta(\bm{\tau}_t, \bm{u}_t \mid \bm{z}_{\hat{t}}; \bm{c}) \bigg)^2\bigg],
    \end{split}
\]
where $\bm{z}_{\hat{t}} \sim f_\phi(\bm{s}_{\hat{t}}; \bm{c})$, $\bm{z}_{\hat{t+1}} \sim f_{\bar{\phi}}(\bm{s}_{\hat{t+1}}; \bm{c})$, and $\bar{\phi}$ denotes the parameters of the target network for the coach's strategy predictor $f$. We build our network on top of A-QMIX but use a separate multi-head attention (MHA) layer to encode the global state that the coach observes. For the mixing network, we also use the coach's output from the MHA layer ($h^{\text{team}}$) for mixing the individual $Q^a$ to form the team $Q^{\text{tot}}$. The entire architecture is illustrated in Figure~\ref{fig:architecture}.
\subsection{Regularizing with variational objective}
Inspired by recent work that applied variational inference to regularize the learning of a latent space in reinforcement learning~\citep{rakelly2019efficient,wang2020multi}, we also introduce a variational objective to stabilize the training. Intuitively, an agent's behavior should be consistent with its assigned strategy. In other words, the received strategy should be identifiable from the agent's future trajectory. Therefore, we propose to maximize the mutual information between the strategy and the agent's future observation-action pairs $\zeta^{a}_t = (o^a_{t+1}, u^a_{t+1}, o^a_{t+2}, u^a_{t+2}, \dots, o^a_{t+T-1}, u^a_{t+T-1})$. We maximize the following variational lower bound:
\begin{equation}
    \label{eq:var}
    % \begin{split}
    % I(z^a_t ; \zeta^a_t, \bm{s}_t) 
    % &= \mathbb{E}_{\bm{s}_t, z^a_t, \zeta^a_t}\bigg[\log\frac{q_\xi(z^a_t | \zeta^a_t, \bm{s}_t)}{p(z^a | \bm{s}_t)}\bigg] + \KL\bigg(p(z^a_t | \zeta^a_t, \bm{s}_t), q_\xi(z^a_t | \zeta^a_t, \bm{s}_t))\bigg)\\
    % &\geq \mathbb{E}_{\bm{s}_t, z^a_t, \zeta^a_t}\bigg[\log\frac{q_\xi(z^a_t | \zeta^a_t, \bm{s}_t)}{p(z^a | \bm{s}_t)}\bigg] = \mathbb{E}_{\bm{s}_t, z^a_t, \zeta^a_t}\bigg[\log q_\xi(z^a_t | \zeta^a_t, \bm{s}_t)\bigg] + H(z^a_t | \bm{s}_t).
    % \end{split}
    \begin{split}
    I(z^a_t ; \zeta^a_t, \bm{s}_t) \geq \mathbb{E}_{\bm{s}_t, z^a_t, \zeta^a_t}\bigg[\log q_\xi(z^a_t | \zeta^a_t, \bm{s}_t)\bigg] + H(z^a_t | \bm{s}_t).
    \end{split}
\end{equation}
We defer the full derivation to Appendix A. Here $H(\cdot)$ denotes the entropy and $q_\xi$ is the variational distribution parameterized by $\xi$. We further adopt the Gaussian factorization for $q_\xi$ as in~\citet{rakelly2019efficient}, i.e. 
\[
q_\xi(z^a_t | \zeta^a_t, \bm{s}_t) \propto q_\xi^{(t)}(z^a_t | \bm{s}_t, u^a_t) \prod_{k=t+1}^{t+T-1} q_\xi^{(k)}(z^a_t | o^a_k, u^a_k),
\]
where each $q_\xi^{(\cdot)}$ is a Gaussian distribution. So $q_\xi$ predicts the $\hat{\mu}^a_t$ and $\hat{\Sigma}^a_t$ of a multivariate normal distribution from which we calculate the log-probability of $z^a_t$. In practice, $z^a_t$ is sampled from $f_\phi$ using the re-parameterization trick~\citep{kingma2013auto}. The objective is  
$
\mathcal{L}_\text{var}(\phi, \xi) = - \lambda_1 \mathbb{E}_{\bm{s}_t, z^a_t, \zeta^a_t}[\log q_\xi(z^a_t | \zeta^a_t, \bm{s}_t)] - \lambda_2 H(z^a_t | \bm{s}_t)
$, where $\lambda_1$ and $\lambda_2$ are tunable coefficients.
\subsection{Reducing the communication frequency}
So far, we have assumed that every $T$ steps the coach broadcasts new strategies for all agents. In practice, broadcasting may incur communication costs. So it is desirable to only distribute strategies when useful. To reduce the communication frequency, we propose an intuitive method that decides whether to distribute new strategies based on the $\ell_2$ distance of the old strategy to the new one. In particular, at time step $t = kT, k \in \mathbb{Z}$, assuming the prior strategy for agent $a$ is $z^a_\text{old}$, the new strategy for agent $a$ is
\vspace{3pt}
\begin{equation}
    \label{eq:broadcast}
    \tilde{z}^a_t = 
    \begin{cases}
        z^a_t \sim f_\phi(\bm{s}, \bm{c}) & \text{if}~||z^a_t - z^a_\text{old}||_2 \geq \beta\\
        z^a_\text{old} & \text{otherwise}.\\
    \end{cases}
\end{equation}

For a general time step $t$, the individual strategy for $a$ is therefore $\tilde{z}^a_{\hat{t}}$. Here $\beta$ is a manually specified threshold. Note that we can train a single model and apply this criterion for all agents. By adjusting $\beta$, one can easily achieve different communication frequencies. Intuitively, when the previous strategy is ``close" to the current one, it should be more tolerable to keep using it. The intuition can be operationalized concretely when the learned $Q^\text{tot}_\theta$ approximates the optimal action-value $Q^\text{tot}_*$ well and has a relatively small Lipschitz constant. Specifically, we have the following theorem:
\vspace{5pt}
\begin{theorem}
    \label{thm:v}
    Denote the optimal action-value and value functions by $Q^\text{tot}_*$ and $V^\text{tot}_*$. Denote the action-value and value functions corresponding to receiving new strategies every time from the coach by $Q^\text{tot}$ and $V^\text{tot}$, and those corresponding to following the strategies distributed according to \eqref{eq:broadcast} as $\tilde{Q}$ and $\tilde{V}$, i.e. $\tilde{V}(\bm{\tau}_t | \tilde{\bm{z}}_{\hat{t}}; \bm{c}) = \max_{\bm{u}} \tilde{Q}(\bm{\tau}_{\hat{t}}, \bm{u} | \tilde{\bm{z}}_t; \bm{c})$. Assume for any trajectory $\bm{\tau}_t$, actions $\bm{u}_t$, current state $\bm{s}_t$, the most recent state the coach distributed strategies $\bm{s}_{\hat{t}}$, and the players' characteristics $\bm{c}$, $||Q^{\text{tot}}(\bm{\tau}_t, \bm{u}_t, f(\bm{s}_{\hat{t}}); \bm{c}) - Q^\text{tot}_*(\bm{s}_t, \bm{u}_t; \bm{c})||_2 \leq \kappa$, and for any strategies $z^a_1, z^a_2$, $|Q^\text{tot}(\bm{\tau}_t, \bm{u}_t | z^a_1, \bm{z}^{-a}; \bm{c}) - Q^\text{tot}(\bm{\tau}_t, \bm{u}_t | z^a_2, \bm{z}^{-a}; \bm{c})| \leq \eta ||z^a_1 - z^a_2||_2$.
    If the used team strategies $\tilde{\bm{z}_t}$ satisfy $\forall a,t,~||\tilde{z}^a_{\hat{t}} - z^a_{\hat{t}}||_2 \leq \beta$, then we have
    \begin{equation}
        ||V^\text{tot}_*(\bm{s}_t; \bm{c}) - \tilde{V}(\bm{\tau}_t | \tilde{\bm{z}}_{\hat{t}}; \bm{c})||_\infty \leq \frac{2(n_a\eta\beta + \kappa)}{1-\gamma},
    \end{equation}
    where $n_a$ is the number of agents and $\gamma$ is the discount factor.
\end{theorem}

We defer the proof to Appendix B. The method described in~\eqref{eq:broadcast} satisfies the condition in Theorem~\ref{thm:v} and therefore when $\beta$ is small, distributing strategies according to~\eqref{eq:broadcast} results in a bounded drop in performance.

%% file: content/experiment.tex
\section{Experiments}
We design the experiments to 1) verify the effectiveness of the coach agent; 2) investigate how performance varies with the interval $T$; 3) test if the variational objective is useful; and 4) understand how much the performance drops by adopting the method specified by~\eqref{eq:broadcast}. We test our idea on a resource collection task built on the multi-agent particle environment~\citep{lowe2017multi}, a multi-agent rescue game, and customized micromanagement tasks in StarCraft.
\begin{figure}[t]
    \centering
    \includegraphics[width=0.93\columnwidth]{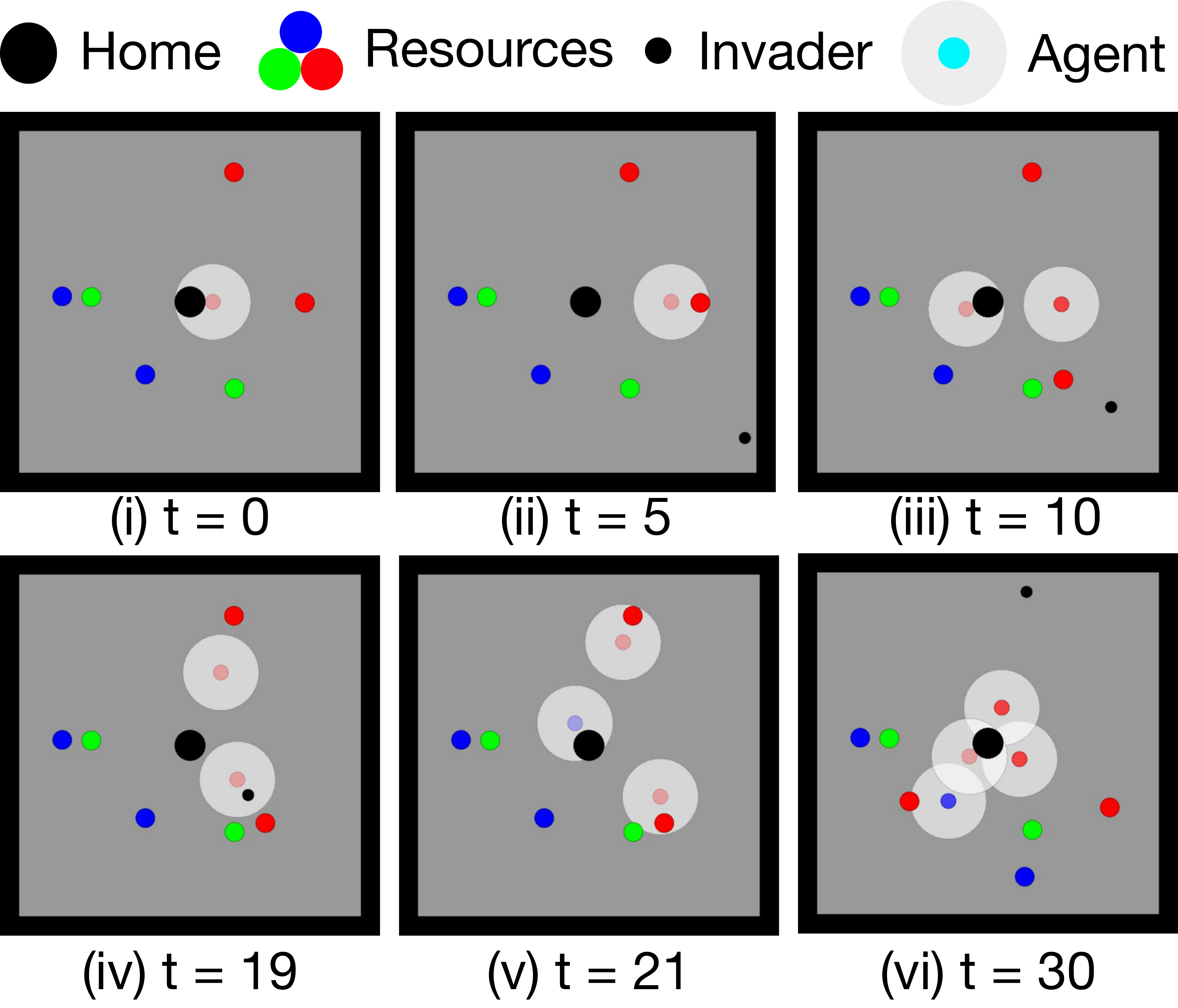}
    \caption{An example testing episode up to $t=30$ with communication interval $T=4$. The team composition changes over time. Here, $c^a$ is represented by rgb values, $c^a=(r,g,b,v)$. For illustration, we set agents rgb to be one-hot but it can vary in practice. (i) an agent starts at home; (ii) the invader (black) appears while the agent (red) goes to the red resource; (iii) another agent is spawned while the old agent brings its resource home; (iv) one agent goes for the invader while the other for a resource; (v-vi) a new agent (blue) is spawned and goes for the blue resource while other agents (red) are bringing resources home.}
    \label{fig:mpe-env}
    \vspace{-10pt}
\end{figure}
\subsection{Resource Collection}
In Resource Collection, a team of agents coordinates to collect different resources spread out on a square map with width $1.8$. There are 4 types of entites: the resources, the agents, the home, and the invader. We assume there are 3 types of resources: ($r$)ed, ($g$)reen and ($b$)lue. In the world, always 6 resources appear with 2 of each type. Each agent has 4 characteristics $(c^a_r, c^a_g, c^a_b, v^a)$, where $c^a_x$ represents how efficiently $a$ collects the resource $x$, and $v$ is the agent's \emph{max} moving speed. The agent's job is to collect as many resources as possible, bring them home, and catch the invader if it appears. If $a$ collects $x$, the team receives a reward of $10\cdot c^a_x$ as reward. Agents can only hold one resource at a time. After collecting it, the agent must bring the resource home before going out to collect more. Bringing a resource home yields a reward of $1$. Occasionally the invader appears and goes directly to the home. Any agent catching the invader will have $4$ reward. If the invader reaches the home, the team is penalized by a $-4$ reward. Each agent has 5 actions: accelerate up / down / left / right and decelerate, and it observes anything within a distance of $0.2$. The maximum episode length is $145$ time steps. In training, we allow scenarios to have 2 to 4 agents, and for each agent, $c^a_r, c^a_g, c^a_b$ are drawn uniformly from $\{0.1, 0.5, 0.9\}$ and the max speed $v^a$ is drawn from $\{0.3, 0.5, 0.7\}$. We design 3 testing tasks: 5-agent task, 6-agent task, and a varying-agent task. For each task, we generate 1000 different scenarios $\bm{c}$. Each scenario includes $n_a$ agents, 6 resources, and an invader. For agents, $c^a_r,c^a_g,c^a_b$ are chosen uniformly from the interval $[0.1, 0.9]$ and $v^a$ from $[0.2, 0.8]$. For a particular scenario in the varying agent task, starting from 4 agents, the environment randomly adds or drops an agent every $\nu$ steps as long as the number of agents remains in $[2, 6]$. $\nu$ is drawn from the uniform distribution $\mathcal{U}(8,12)$. See Figure~\ref{fig:mpe-env} for an example run of the learned policy.

\begin{figure*}[t]
    \centering
    \includegraphics[width=0.9\textwidth]{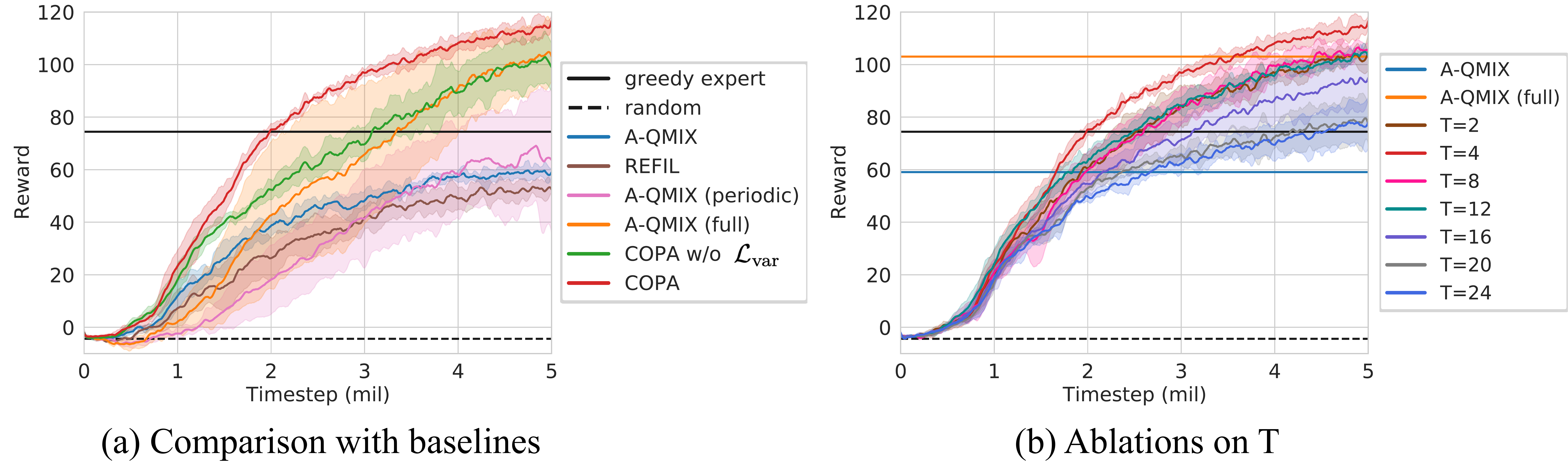}
    \caption{Training curves for Resource Collection. (a) comparison against A-QMIX, REFIL and COPA without the variational objective. Here we choose $T=4$; (b) ablations on the communication interval $T$. All results are averaged over 5 seeds.}
    \label{fig:mpe-exp}
    \vspace{-5pt}
\end{figure*}
\vspace{-4pt}
\paragraph{Effectiveness of Coach } Figure~\ref{fig:mpe-exp}(a) shows the training curve when the communication interval is set to $T=4$. The black solid line is a hand-coded greedy algorithm where agents always go for the resource they are best at collecting, and whenever the invader appears, the closest agent goes for it. We see that without global information, A-QMIX and REFIL are significantly worse than the hand-coded baseline. Without the coach, we let all agents have the global view every $T$ steps in A-QMIX (periodic) but it barely improves over A-QMIX. A-QMIX (full) is fully observable, i.e., all agents have global view. Without the variational objective, COPA performs comparably against A-QMIX (full). With the variational objective, it becomes even better than A-QMIX (full). The results confirm the importance of global coordination and the coach-player hierarchy.
\vspace{-4pt}
\paragraph{Communication Interval} To investigate how performance varies with $T$, we train with different $T$ chosen from $[2,4,8,12,16,20,24]$ in Figure~\ref{fig:mpe-exp}(b). The performance peaks at $T=4$, countering the intuition that smaller $T$ is better. This result suggests that the coach is most useful when it can cause the agents to behave smoothly/consistently over time.
\vspace{-4pt}
\paragraph{Zero-shot Generalization} 
We apply the learned model with $T=4$ to the 3 testing environments. Results are provided in Table~\ref{tab:mpe-test}. The communication frequency is calculated according to the fully centralized setting. For instance, when $T=4$ and $\beta=0$, it results in an average $25\%$ centralization frequency.

\begin{table}[h]
\centering
\resizebox{1.0\columnwidth}{!}{%
  \begin{tabular}{lcccc}
    \toprule
    Method &  Env. ($n=5$) & Env. ($n=6$) & Env. (varying $n$) & $f$ \\
    \midrule
    Random Policy       &   6.9                         &  10.4                         &   2.3                         & N/A  \\
    Greedy Expert       & 115.3                         & 142.4                         &  71.6                         & N/A  \\
    REFIL               &  90.5{\footnotesize$\pm$1.5}  & 109.3{\footnotesize$\pm$1.6}  &  61.5{\footnotesize$\pm$0.9}  &  0  \\
    A-QMIX              &  96.9{\footnotesize$\pm$2.1}  & 115.1{\footnotesize$\pm$2.1}  &  66.2{\footnotesize$\pm$1.6}  &  0  \\
    A-QMIX (periodic)   &  93.1{\footnotesize$\pm$20.4} & 104.2{\footnotesize$\pm$22.6} &  68.9{\footnotesize$\pm$12.6} & 0.25 \\
    A-QMIX (full)       & 157.4{\footnotesize$\pm$8.5}  & 179.6{\footnotesize$\pm$9.8}  & 114.3{\footnotesize$\pm$6.2}  &  1  \\
    \midrule
    COPA ($\beta=0$)    & 175.6{\footnotesize$\pm$1.9}  & 203.2{\footnotesize$\pm$2.5}  & 124.9{\footnotesize$\pm$0.9}  & 0.25 \\
    COPA ($\beta=2$)    & 174.4{\footnotesize$\pm$1.7}  & 200.3{\footnotesize$\pm$1.6}  & 122.8{\footnotesize$\pm$1.5}  & 0.18 \\
    COPA ($\beta=3$)    & 168.8{\footnotesize$\pm$1.7}  & 195.4{\footnotesize$\pm$1.8}  & 120.0{\footnotesize$\pm$1.6}  & 0.13 \\
    COPA ($\beta=5$)    & 149.3{\footnotesize$\pm$1.4}  & 174.7{\footnotesize$\pm$1.7}  & 104.7{\footnotesize$\pm$1.6}  & 0.08 \\
    COPA ($\beta=8$)    & 109.4{\footnotesize$\pm$3.6}  & 130.6{\footnotesize$\pm$4.0}  &  80.6{\footnotesize$\pm$2.0}  & 0.04 \\
    \bottomrule
  \end{tabular}
 }
\caption{Mean episodic reward on unseen environments with more agents and \emph{dynamic team composition}. Results are computed from 5 models trained with different seeds. Communication frequency ($f$) is compared to communicating with all agents at every step.}
\label{tab:mpe-test}
\end{table}

\begin{wrapfigure}{r}{0.53\columnwidth}
    \centering
    \vspace{-5pt}
    \includegraphics[width=0.53\columnwidth]{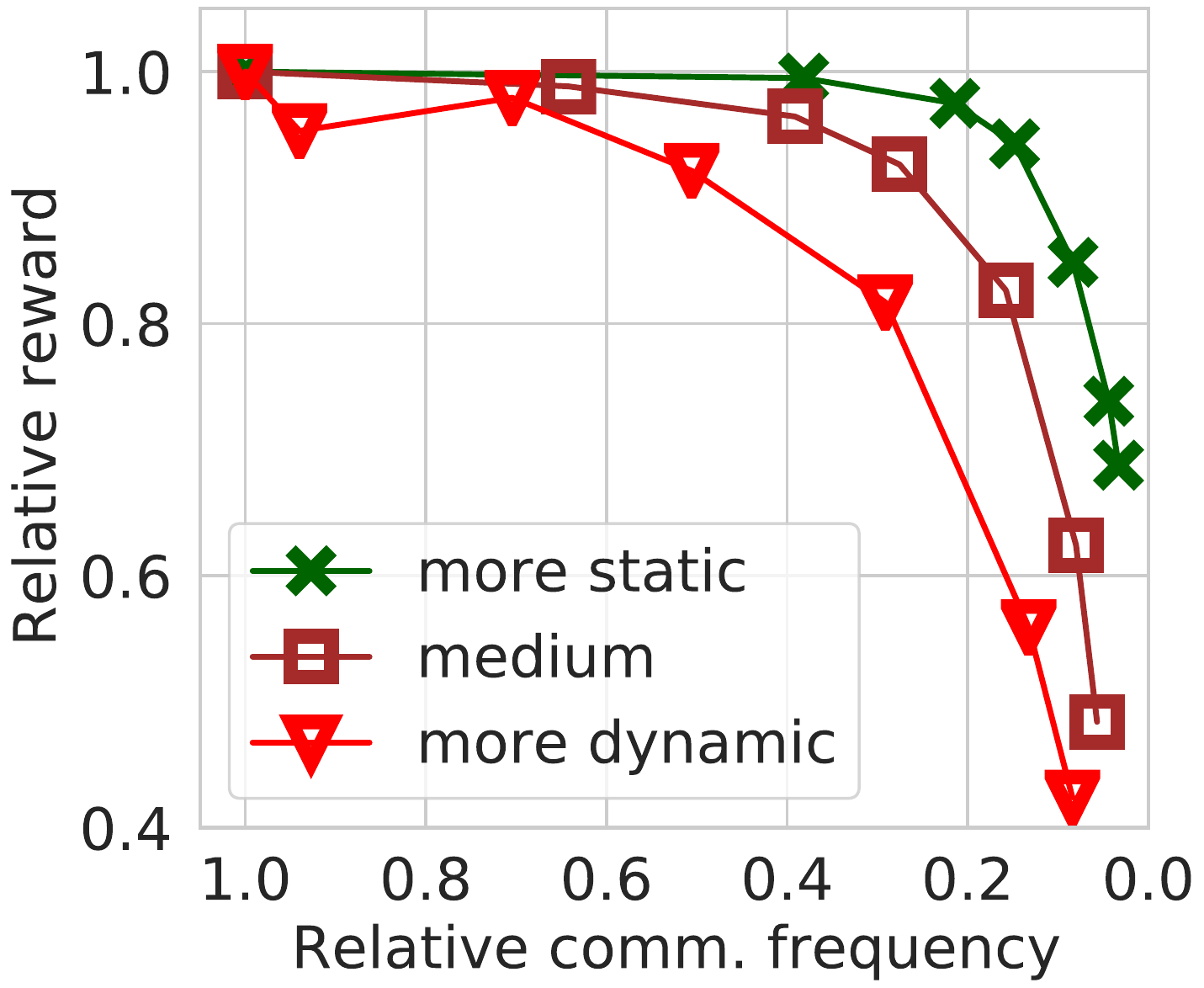}
    \vspace{-15pt}
    \caption{Sensitivity to communication frequency $f$.}
    \label{fig:mpe-diff}
    \vspace{-10pt}
\end{wrapfigure}

As we increase $\beta$ to suppress the distribution of strategies, we see that the performance shows no significant drop until $13\%$ centralization frequency. Moreover, we apply the same model to 3 environments that are dynamic to different extents (see Figure~\ref{fig:mpe-diff}).
In the more static environment, resources are always spawned at the same locations. In the medium environment, resources are spawned randomly but there is no invader.
The more dynamic environment is the 3rd environment in Table~\ref{tab:mpe-test} where the team is dynamic in composition and there exists the invader.
In Figure~\ref{fig:mpe-diff}, the x-axis is normalized according to the communication frequency $f$ when $\beta=0$, and the y-axis is normalized by the corresponding performance. As expected, performance drops faster in more dynamic environments as we decrease $f$.

\subsection{Rescue Game}
\begin{figure}[b]
    \centering
    \vspace{-8pt}
    \includegraphics[width=\columnwidth]{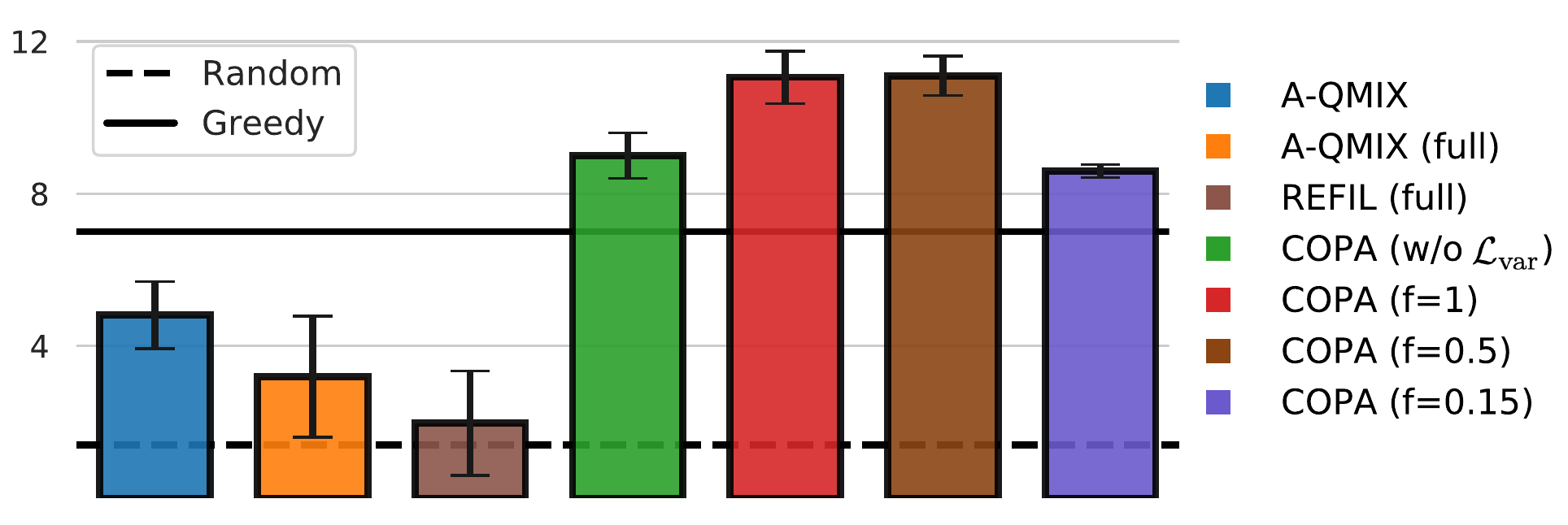}
    \caption{Mean episodic reward over $500$ unseen Rescue games on 3 models trained with 3 seeds. COPA $(f=x)$ denotes COPA with communication frequency $x$. Greedy algorithm matches the $k$-th skillful agent for the $k$-th emergent building.}
    \label{fig:rescue}
\end{figure}
Search-and-rescue is a natural application for multi-agent systems. In this section we apply COPA to different rescue games. In particular, we consider a $10\times10$ grid-world, where each grid contains a building. At any time step, each building is subject to catch a fire. When a building $b$ is on fire, it has an emergency level $c^b \sim \mathcal{U}(0, 1)$. Within the world, at most $10$ buildings will be on fire at the same time. To protect the buildings, we have $n$ ($n$ is a random number from $2$ to $8$) robots who are the surveillance firefighters. Each robot $a$ has a skill-level $c^a \in [0.2, 1.0]$. A robot has 5 actions, moving up/down/left/right and put out the fire. If $a$ is at a building on fire and chooses to put out the fire, the emergency level will be reduced to $c^b \leftarrow \max(c^b - c^a, 0)$. At each time step $t$, the overall-emergency is given by $c^B_t = \sum_b (c^b)^2$ since we want to penalize greater spread of the fire. The reward is defined as $r_t = c^B_{t-1} - c^B_{t}$, the amount of spread that the team prevents. During training, we sample $n$ from $3-5$ and these robots are spawned randomly across the world. Each agent's skill-level is sampled from $[0.2, 0.5, 1.0]$. Then a random number of $3-6$ buildings will catch a fire. During testing, we enlarge $n$ to $2-8$ agents and sample up to $8$ buildings on fire. We summarize the results in the Figure~\ref{fig:rescue}.

Interestingly, we find that A-QMIX/REFIL with full views perform worse than A-QMIX with partial views. We conjecture this is because the full view complicates learning.
\begin{wrapfigure}{r}{0.53\columnwidth}
    \centering
    \vspace{-5pt}
    \includegraphics[width=0.53\columnwidth]{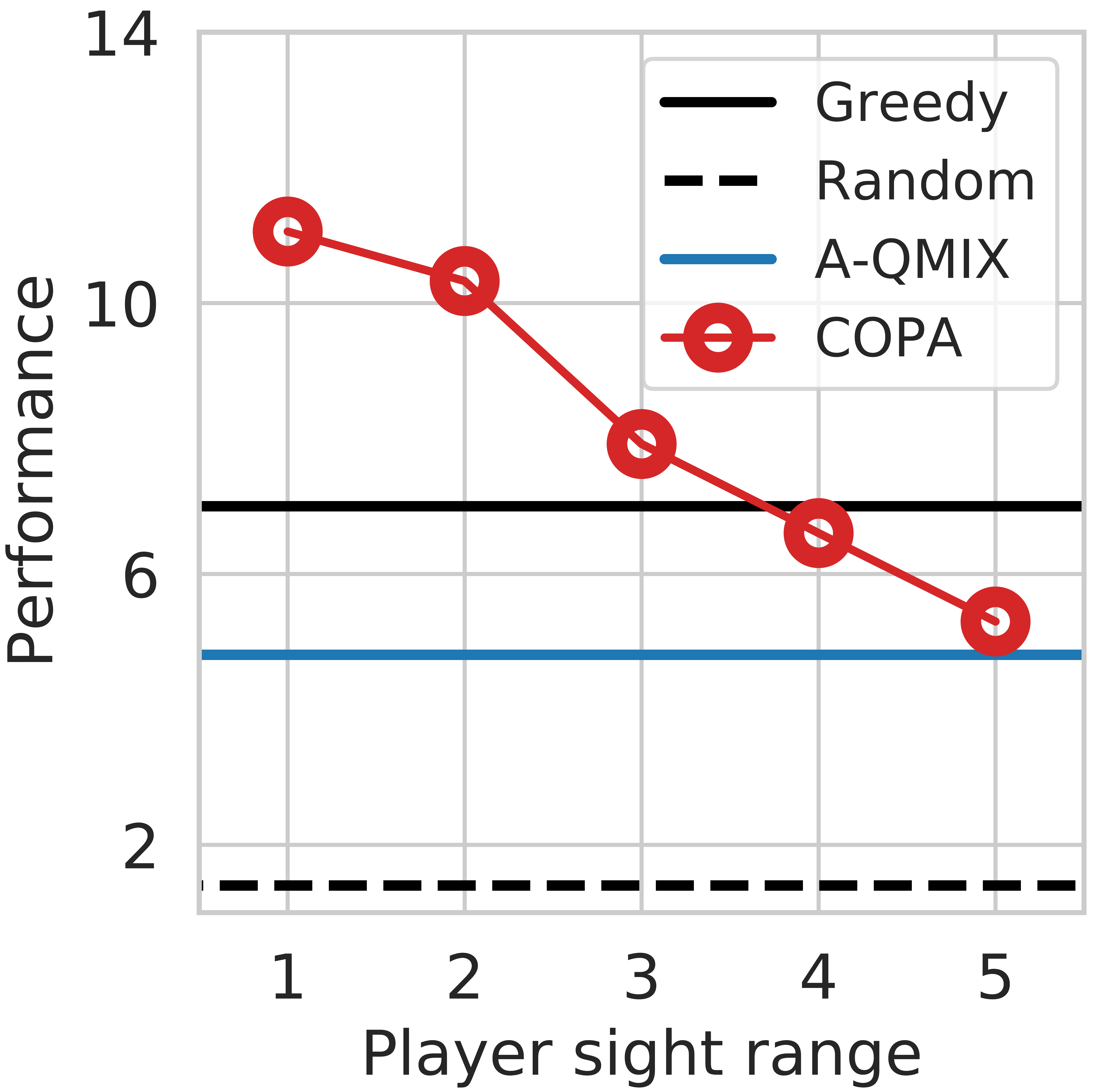}
    \vspace{-20pt}
    \caption{Performance with different sight ranges of the players.}
    \label{fig:rescue_sight}
    \vspace{-10pt}
\end{wrapfigure}
On the other hand, COPA consistently outperforms all baselines even with a communication frequency as low as $0.15$. In addition, we conduct an ablation on the sight range of the players in Figure~\ref{fig:rescue_sight}.
As we increase the sight range of the players from 1 to 5, where 5 is equivalent to granting full views to all players, the performance on test scenarios drops. This result raises the possibility that agents and players might benefit from having different views of the environment.

\subsection{StarCraft Micromanagement Tasks}
\begin{figure}[t]
    \centering
    \includegraphics[width=\columnwidth]{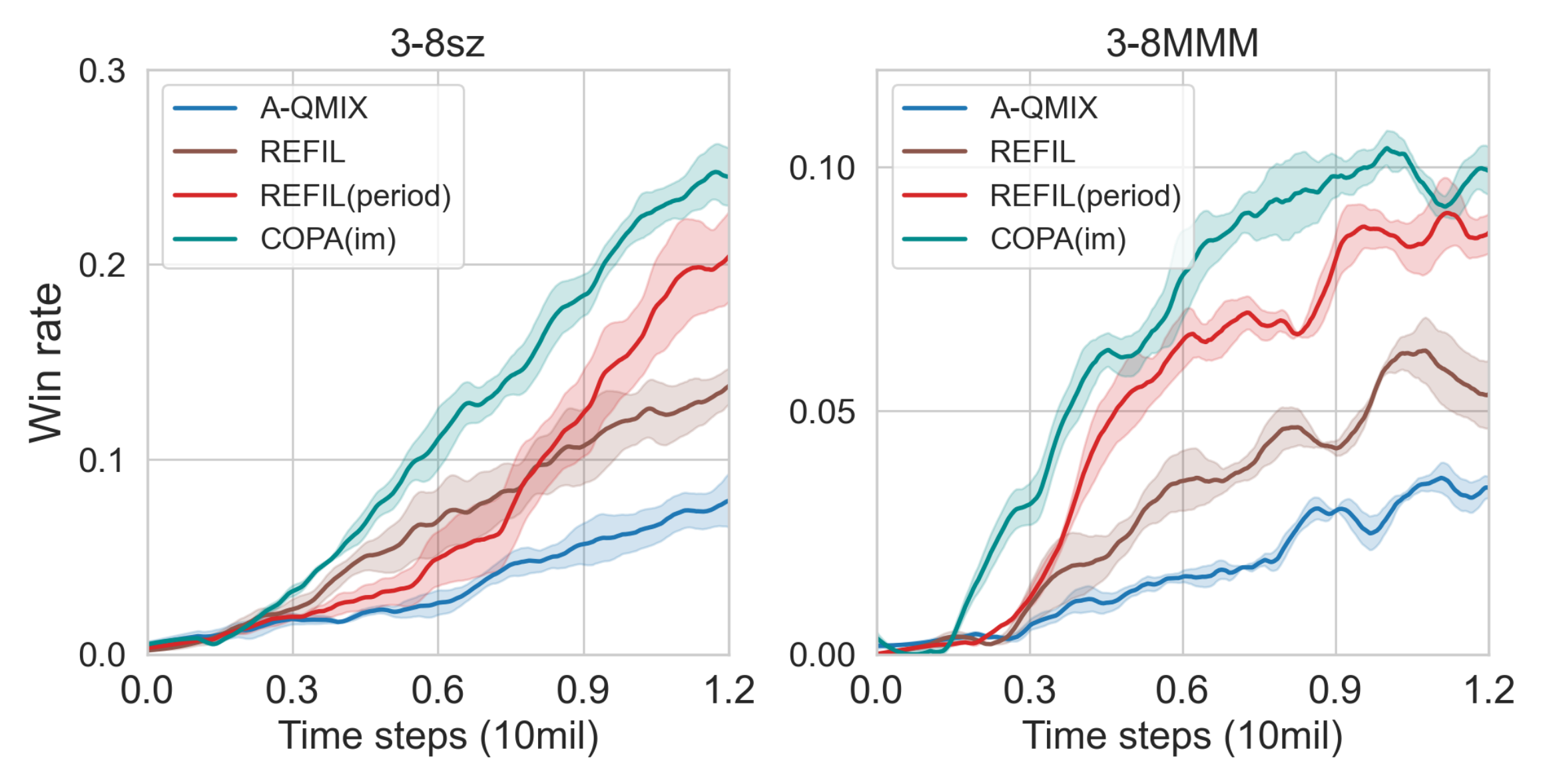}
    \caption{Training curves of the win rate on the 3-8sz and 3-8MMM environments. COPA(im) indicates COPA applied with the additional imaginary grouping objective as in REFIL \cite{iqbal2020ai}. Each experiment is conducted over 3 independent runs. The shaded area indicates the standard error of the mean.}
    \label{fig:smac}
\end{figure}
Finally, we apply COPA on the more challenging StarCraft multi-agent challenge (SMAC)~\cite{samvelyan2019starcraft}. SMAC involves controlling a team of units to defend a team of enemy units. We test on the two settings, 3-8sz and 3-8MMM, from~\citet{iqbal2020ai}. Here, 3-8sz means the scenario consists of symmetric teams of between 3 to 8 units, with each agent being either a Zealot or a Stalker. Similarly 3-8MMM has between 0 and 2 Medics and 3 to 6 Marines/Marauders. In \citet{iqbal2020ai}, both teams are grouped together and all agents can fully observe their teammates. To make the task more challenging, we modified both environments such that the agent team is randomly divided into 2 to 4 groups while the enemy team is organized into 1 or 2 groups. Each group is spawned together at a random place. Specifically, the center of any enemy group is drawn uniformly on the boundary of a circle of radius $7$, while the center of any group of the controlled agents is drawn uniformly from within the circle of radius $6$. Then we grant each unit a sight range of $3$. The learning curves are shown in Figure~\ref{fig:smac}. 

As we find that on SMAC environments, the imaginary objective proposed in ~\cite{iqbal2020ai} further improves the performance of A-QMIX, we also incorporate the imaginary objective for COPA, denoted by COPA(im). Then we compare COPA(im) with A-QMIX, REFIL, and REFIL with the omniscient view at the same maximum frequency (e.g. $T=4$) that COPA distributes the strategies, denoted by REFIL(periodic). COPA outperforms A-QMIX and REFIL without the omniscient views by a large margin. Moreover, COPA performs slightly better than REFIL (periodic). But note that COPA can dynamically control the communication frequency which REFIL(periodic) cannot.
We further provide the results under different communication frequencies in Table~\ref{tab:smac}. COPA's performance remains strong with the communication frequency as low as $0.04$ and $0.10$ respectively on 3-8sz and 3-8MMM environments.

% \begin{table}[h]
% \centering
% \resizebox{1.0\columnwidth}{!}{%
%   \begin{tabular}{lcccc}
%     \toprule
%     \multirow{2}{*}{Method} &  \multicolumn{2}{c}{3-8sz} & \multicolumn{2}{c}{3-8MMM} \\
%                             & Win Rate \% & $f$ & Win Rate \% & $f$ \\
%     \midrule
%     REFIL   &  19.83{\footnotesize$\pm$0.93}  & 0 &  7.79{\footnotesize$\pm$0.47}  &  0  \\
%     \midrule
%     COPA ($\beta=0$)    & 36.09{\footnotesize$\pm$2.11}  & 0.25  & 11.30{\footnotesize$\pm$1.48}  & 0.25 \\
%     COPA ($\beta=2$)    & 35.14{\footnotesize$\pm$1.40}  & 0.08  & 10.55{\footnotesize$\pm$0.37}  & 0.15 \\
%     COPA ($\beta=3$)    & 27.49{\footnotesize$\pm$1.40}  & 0.05  & 12.22{\footnotesize$\pm$0.86}  & 0.12 \\
%     COPA ($\beta=4$)    & 16.26{\footnotesize$\pm$1.12}  & 0.02  & 8.73{\footnotesize$\pm$0.64}  & 0.08 \\
%     \bottomrule
%   \end{tabular}
%  }
% \caption{Performance on 3-8sz and 3-8MMM versus the communication frequency ($f$) controlled by $\beta$. Results are computed from 3 models on 500 test episodes.}
% \label{tab:smac}
% \vspace{-5pt}
% \end{table}

\begin{table}[h]
\centering
\resizebox{1.0\columnwidth}{!}{%
  \begin{tabular}{lcccc}
    \toprule
    \multirow{2}{*}{Method} &  \multicolumn{2}{c}{3-8sz} & \multicolumn{2}{c}{3-8MMM} \\
                            & Win Rate \% & $f$ & Win Rate \% & $f$ \\
    \midrule
    REFIL   &  12.90{\footnotesize$\pm$3.36}  & 0 &  5.78{\footnotesize$\pm$0.23}  &  0  \\
    REFIL (periodic) & 22.11{\footnotesize$\pm$8.04} & 0.25 & 8.53{\footnotesize$\pm$0.99} & 0.25 \\
    \midrule
    COPA ($\beta=0$)    & 26.07{\footnotesize$\pm$3.03}  & 0.25  & 11.62{\footnotesize$\pm$1.23}  & 0.25 \\
    COPA ($\beta=1$)    & 26.61{\footnotesize$\pm$2.73}  & 0.16  & 10.42{\footnotesize$\pm$1.63}  & 0.20 \\
    COPA ($\beta=2$)    & 22.65{\footnotesize$\pm$1.34}  & 0.09  & 10.35{\footnotesize$\pm$0.84}  & 0.15 \\
    COPA ($\beta=3$)    & 14.25{\footnotesize$\pm$1.11}  & 0.04  & 9.01{\footnotesize$\pm$1.17}  & 0.10 \\
    \bottomrule
  \end{tabular}
 }
\caption{Performance on 3-8sz and 3-8MMM versus the communication frequency ($f$) controlled by $\beta$. Results are computed from 3 models on 500 test episodes.}
\label{tab:smac}
\end{table}

%% file: content/related.tex
\section{Related Work}
In addition to the related work highlighted in Section 2, here we review other relevant works in four categories.
\vspace{-5pt}
\paragraph{Centralized Training with Decentralized Execution} Centralized training with decentralized execution (CTDE) assumes agents execute independently but uses the global information for training. A branch of methods investigates factorizable $Q$ functions~\citep{sunehag2017value,rashid2018qmix,mahajan2019maven,son2019qtran} where the team $Q$ is decomposed into individual utility functions. Some other methods adopt the actor-critic method where only the critic is centralized~\citep{foerster2017counterfactual,lowe2017multi}. However, most CTDE methods by structure require fixed-size teams and are often applied to homogeneous teams.
\vspace{-10pt}
\paragraph{Methods for Dynamic Team Compositions} Several recent works in transfer learning and curriculum learning attempt to transfer policy of small teams for larger teams~\citep{carion2019structured,shu2019m3rl,agarwal2019learning,wang2020few,long2020evolutionary}. These works mostly consider teams with different numbers of homogeneous agents. Concurrently, \citet{iqbal2020ai} and \citet{zhang2020multi} also consider using the attention mechanism to deal with a variable number of agents. But both methods are fully decentralized and we focus on studying how to incorporate global information for ad hoc teams when useful.
\vspace{-5pt}
\paragraph{Ad Hoc Teamwork and Networked Agents} Standard ad hoc teamwork research mainly focuses on the single ad hoc agent and assumes no control over the teammates~\citep{PAIR11-katie,AAMAS12-Barrett}. Recently, \citet{grizou2016collaboration} and \citet{mirsky2020penny} consider communication between ad hoc agents but assume the communication protocol is pre-defined. Decentralized networked agents assume information can propagate among agents and their neighbors~\citep{kar2013cal,macua2014distributed,foerster2016learning,suttle2019multi,zhang2018fully}. However, to our knowledge, networked agents have not been applied to teams with dynamic compositions yet.
\vspace{-5pt}
\paragraph{Hierarchical Reinforcement Learning} Hierarchical RL/MARL decomposes the task into hierarchies: a meta-controller selects either a temporal abstracted action~\citep{bacon2017option}, called an option, or a goal state~\citep{vezhnevets2017feudal} for the base agents. Then the base agents shift their purposes to finish the assigned option or reach the goal. Therefore usually the base agents have different learning objective from the meta-controller. Recent deep MARL methods also demonstrate role emergence~\citep{wang2020multi} or skill emergence~\citep{yang2019hierarchical}. But the inferred role/skill is only conditioned on the individual trajectory. The coach in our method uses global information to determine the strategies for the base agents. To our knowledge, we are the first to apply such a hierarchy for teams with varying numbers of heterogeneous agents.

%% file: content/conclusion.tex
\vspace{-5pt}
\section{Conclusion}
We investigated a new setting of multi-agent reinforcement learning problems, where both the team size and members' capabilities are subject to change. To this end, we proposed a coach-player framework, COPA, where the coach coordinates with a global view but players execute independently with local views and the coach's strategy. We developed a variational objective to regularize the learning of the strategies and introduces an intuitive method to suppress unnecessary distribution of strategies. Results on three different environments demonstrate that the coach is important for learning when the team composition might change. COPA achieves strong zero-shot generalization performance with relatively low communication frequency. Interesting future directions include investigating better communication strategies between the coach and the players and introducing multiple coach agents with varying levels of views.

%% file: content/appendix.tex
\onecolumn
\section{Regularizing with variational objective}
\label{apx:var-reg}
We provide the full derivation of Equation~\eqref{eq:var} in the following:
\begin{equation}
    \label{eq:var-derive}
    \begin{split}
    I(z^a_t ; \zeta^a_t, \bm{s}_t) 
    &= \mathbb{E}_{\bm{s}_t, z^a_t, \zeta^a_t}\bigg[\log\frac{p(z^a_t | \zeta^a_t, \bm{s}_t)}{p(z^a | \bm{s}_t)}\bigg] \ant{by the definition of mutual information} \\
    &= \mathbb{E}_{\bm{s}_t, z^a_t, \zeta^a_t}\bigg[\log\frac{q_\xi(z^a_t | \zeta^a_t, \bm{s}_t)}{p(z^a | \bm{s}_t)}\bigg] + \KL\bigg(p(z^a_t | \zeta^a_t, \bm{s}_t), q_\xi(z^a_t | \zeta^a_t, \bm{s}_t))\bigg)\\
    &\geq \mathbb{E}_{\bm{s}_t, z^a_t, \zeta^a_t}\bigg[\log\frac{q_\xi(z^a_t | \zeta^a_t, \bm{s}_t)}{p(z^a | \bm{s}_t)}\bigg] \ant{since $\KL(\cdot, \cdot) \geq 0$}\\
    &= \mathbb{E}_{\bm{s}_t, z^a_t, \zeta^a_t}\bigg[\log q_\xi(z^a_t | \zeta^a_t, \bm{s}_t)\bigg] + H(z^a_t | \bm{s}_t).
    \end{split}
\end{equation}
\section{Proof of Theorem 1}
\label{apx:thm1proof}
In this section we provide the proof for Theorem~\ref{thm:v}.
\begin{customthm}{1}
    Denote the optimal action-value and value functions by $Q^\text{tot}_*$ and $V^\text{tot}_*$. Denote the action-value and value functions corresponding to receiving new strategies every time from the coach by $Q^\text{tot}$ and $V^\text{tot}$, and thos corresponding to following the strategies distributed according to \eqref{eq:broadcast} as $\tilde{Q}$ and $\tilde{V}$, i.e. $\tilde{V}(\bm{\tau}_t | \tilde{\bm{z}}_{\hat{t}}; \bm{c}) = \max_{\bm{u}} \tilde{Q}(\bm{\tau}_{\hat{t}}, \bm{u} | \tilde{\bm{z}}_t; \bm{c})$. Assume for any trajectory $\bm{\tau}_t$, actions $\bm{u}_t$, current state $\bm{s}_t$, the most recent state the coach distribute strategies $\bm{s}_{\hat{t}}$, and the players' characteristics $\bm{c}$, $||Q^{\text{tot}}(\bm{\tau}_t, \bm{u}_t, f(\bm{s}_{\hat{t}}); \bm{c}) - Q^\text{tot}_*(\bm{s}_t, \bm{u}_t; \bm{c})||_2 \leq \kappa$, and for any strategies $z^a_1, z^a_2$, $|Q^\text{tot}(\bm{\tau}_t, \bm{u}_t | z^a_1, \bm{z}^{-a}; \bm{c}) - Q^\text{tot}(\bm{\tau}_t, \bm{u}_t | z^a_2, \bm{z}^{-a}; \bm{c})| \leq \eta ||z^a_1 - z^a_2||_2$.
    If the used team strategies $\tilde{\bm{z}_t}$ satisfies $\forall a,t,~||\tilde{z}^a_{\hat{t}} - z^a_{\hat{t}}||_2 \leq \beta$, then we have
    \begin{equation}
        ||V^\text{tot}_*(\bm{s}_t; \bm{c}) - \tilde{V}(\bm{\tau}_t | \tilde{\bm{z}}_{\hat{t}}; \bm{c})||_\infty \leq \frac{2(n_a\eta\beta + \kappa)}{1-\gamma},
    \end{equation}
    where $n_a$ is the number of agents and $\gamma$ is the discount factor.
    \vspace{-5pt}
\end{customthm}
To summarize, the assumptions assume that the learned action-value function $Q^\text{tot}$ approximates the optimal $Q_*^\text{tot}$ well and has bounded Lipschitz constant with respect to individual action-value functions. Moreover, we assume the individual action-value functions also have bounded Lipschitz constant with respect to the strategies.
\begin{proof}
According to Assumption 2, if $||\tilde{z}^a_t - z^a_{\hat{t}}||_2 \leq \beta$ for all $a$, then
\begin{equation}
    |Q^{\text{tot}}(\bm{\tau}_t, \bm{u}_t | \bm{\tilde{z}}_t, \bm{c}) - Q^{\text{tot}}(\bm{\tau}_t, \bm{u}_t | \bm{z}_{\hat{t}}, \bm{c})| \leq \sum_{a_i, 1\leq i \leq n_a} \eta_1 \eta_2 ||\tilde{z^a_t} - z^a_{\hat{t}}||_2 \leq n_a \eta_1 \eta_2 \beta.
\end{equation}
For notation convenience, we ignore the superscript of $\textit{tot}$ and the condition on $\bm{c}$. For a state $\bm{s}$, denote the action the learned policy take as $\bm{u}^\dagger$, i.e. $\bm{u}^\dagger \triangleq \argmax_{\bm{u}} Q(\bm{\tau}, \bm{u})$. Similarly we can define $\bm{u}^*$ and $\bm{\tilde{u}}$ as the action one would take according to the optimal $Q_*$ and the action-value $\tilde{Q}$ estimated using the old strategy. From Assumption 1, we know that
\begin{equation}
    Q_*(\bm{s}, \bm{u}^\dagger) \geq Q(\bm{\tau}, \bm{u}^\dagger) - \kappa \geq Q(\bm{\tau}, \bm{u}^*) - \kappa \geq Q_*(\bm{s}, \bm{u}^*) - 2\kappa.
\end{equation}
Therefore taking $\bm{u}^\dagger$ will result in at most $2\kappa$ performance drop at this single step. Similarly, denote $\epsilon_0 = n_a\eta_1\eta_2\beta$, then
\begin{equation}
    Q(\bm{\tau}, \bm{\tilde{u}}) \geq \tilde{Q}(\bm{\tau}, \bm{\tilde{u}}) - \epsilon_0 \geq \tilde{Q}(\bm{\tau}, \bm{u}^\dagger) - \epsilon_0 \geq Q(\bm{\tau}, \bm{u}^\dagger) - 2\epsilon_0.
\end{equation}
Hence $Q_*(\bm{s},\bm{\tilde{u}})  \geq Q_*(\bm{s}, \bm{u}^*) - 2(\epsilon_0 + \kappa)$. Note that this means taking the action $\tilde{\bm{u}}$ in the place of $\bm{u}^*$ at state $\bm{s}$ will result in at most $2(\epsilon_0 + \kappa)$ performance drop. This conclusion generalizes to any step $t$. Therefore, if at each single step the performance is bounded within $2(\epsilon_0 + \kappa)$, then overall the performance is within $2(\epsilon_0 + \kappa)/(1 - \gamma)$.
\end{proof}
\section{Training Details}
\label{apx:training}
For both Resource Collection and Rescue Game, we set the max total number of training steps to 5 million. Then we use the exponentially decayed $\epsilon$-greedy algorithm as our exploration policy, starting from $\epsilon_0 = 1.0$ to $\epsilon_n = 0.05$. We parallelize the environment with 8 threads for training. Experiments are run on the GeForce RTX 2080 GPUs. We provide the algorithm hyper-parameters in Table~\ref{tab:hyper}.
\begin{table}[h]
    \centering
    \begin{tabular}{llc}
    \toprule
    Name & Description & Value \\
    \midrule
      $|\mathcal{D}|$   &  replay buffer size & 100000 \\
      $n_\text{head}$   & number of heads in multi-head attention & 4\\
      $n_\text{thread}$ & number of parallel threads for running the environment & 8\\
      \midrule
      $dh$ & the hidden dimension of all modules & 128\\
      $\gamma$ & the discount factor & 0.99 \\
      $lr$ & learning rate & 0.0003 \\
      & optimizer & RMSprop \\
      $\alpha$ & $\alpha$ value in RMSprop & 0.99\\
      $\epsilon$ & $\epsilon$ value in RMSprop & 0.00001 \\
      $n_\text{batch}$ & batch size & 256 \\
      grad clip & clipping value of gradient & 10\\
      target update frequency & how frequent do we update the target network & 200 updates \\
      $\lambda_1$ & $\lambda_1$ in variational objective & 0.001\\
      $\lambda_2$ & $\lambda_2$ in variational objective & 0.0001\\
      \bottomrule
    \end{tabular}
    \caption{Hyper-parameters in Resource Collection and Rescue Game.}
    \label{tab:hyper}
\end{table}

For StarCraft Micromanagement, we follow the same setup from \citep{iqbal2020ai} and train all methods on the 3-8sz and 3-8MMM maps for 12 millions steps. To regularize the learning, we use $\lambda_1 = 0.00005$ and $\lambda_2 = 0.000005$ for both maps. For all experiments, we set the default period before centralization to $T=4$.